\documentclass[letterpaper, 10 pt, conference]{ieeeconf}  

\usepackage{graphics}
\usepackage{graphicx}
\usepackage{cite}
\usepackage{subfigure}
\usepackage[T1]{fontenc}

\usepackage{gensymb}

\IEEEoverridecommandlockouts

\title{\LARGE \bf
 Shoulder abduction loading affects motor coordination in individuals with chronic stroke, informing targeted rehabilitation}

\author{Aleksandra Kalinowska$^{1}$, Kyra Rudy$^{1*}$, Millicent Schlafly$^{1*}$, Kathleen Fitzsimons$^{1}$ \\
    Julius P Dewald$^{2}$, and Todd D Murphey$^{3}$
\thanks{$^{1}$Aleksandra Kalinowska, Kyra Rudy, Millicent Schlafly, and Kathleen Fitzsimons are with the Department of Mechanical Engineering, Northwestern University, Evanston, IL
        {\tt\small \{ola,kyra,}   {\tt\small milli,k-fitzsimons\}@u.northwestern.edu}}%
\thanks{$^{2}$Todd D Murphey is with the Department of Mechanical Engineering, Northwestern University, Evanston, IL, and with the Department of Physical Therapy and Human Movement Science, Northwestern University, Chicago, IL
        {\tt\small t-murphey@northwestern.edu}}%
\thanks{$^{3}$Julius P Dewald is with the Department of Physical Therapy and Human Movement Science, the Department of Physical Medicine, and the Department of Biomedical Engineering, Northwestern University, Chicago, IL
        {\tt\small j-dewald@northwestern.edu}}%
\thanks{$^{*}$These authors contributed equally.}%
}

\begin{document}

\maketitle
\thispagestyle{empty}
\pagestyle{empty}

\begin{abstract}
Individuals post stroke experience motor impairments, such as loss of independent joint control, weakness, and delayed movement initiation, leading to an overall reduction in arm function. Their motion becomes slower and more discoordinated, making it difficult to complete timing-sensitive tasks, such as balancing a glass of water or carrying a bowl with a ball inside it. Understanding how the stroke-induced motor impairments interact with each other can help design assisted training regimens for improved recovery. In this study, we investigate the effects of abnormal joint coupling patterns induced by flexion synergy on timing-sensitive motor coordination in the paretic upper limb. We design a virtual ball-in-bowl task that requires fast movements for optimal performance and implement it on a robotic system, capable of providing varying levels of abduction loading at the shoulder. We recruit 12 participants (6 individuals with chronic stroke and 6 unimpaired controls) and assess their skill at the task at 3 levels of loading, defined by the vertical force applied at the robot end-effector. Our results show that, for individuals with stroke, loading has a significant effect on their ability to generate quick coordinated motion. With increases in loading, their overall task performance decreases and they are less able to compensate for ball dynamics---frequency analysis of their motion indicates that abduction loading weakens their ability to generate movements at the resonant frequency of the dynamic task. This effect is likely due to an increased reliance on lower resolution indirect motor pathways in individuals post stroke. Given the inter-dependency of loading and dynamic task performance, we can create targeted robot-aided training protocols focused on improving timing-sensitive motor control, similar to existing progressive loading therapies, which have shown efficacy for expanding reachable workspace post stroke. 
\end{abstract}

\section{INTRODUCTION}

Individuals post stroke experience long-term motor impairments, such as weakness or paresis~\cite{beer2007weakness} as well as pathological muscle co-activation patterns~\cite{dewald1995abnormal} that in turn lead to abnormal joint couplings~\cite{Brunnstrom1970, dewald2001abnormal}. Moreover, individuals with stroke rely on an increased use of indirect motor pathways (i.e., corticoreticulospinal tracts) from the non-lesioned hemisphere for movement of paretic limbs~\cite{mcpherson2018progressive}. These underlying neuromuscular changes can be observed through a reduced upper extremity workspace area, overall slower more discoordinated motion, and a loss of independent joint control---especially visible in the abnormal coupling between shoulder abduction and elbow/wrist/finger flexion, referred to broadly as the flexion synergy. Being able to independently assess the stroke-induced motor impairments and to understand how they impact recovery is imperative for designing successful robot-aided rehabilitation. In our work, we are particularly interested in assessing how timing-sensitive motor coordination during dynamic tasks is affected post hemiparetic stroke, and how it interacts with previously quantified motor impairments such as the expression of flexion synergy. 

Although there is significant evidence in literature indicating that timing-sensitive motor coordination is impacted by stroke, there is little consensus as to what underlying mechanisms cause these changes. For one, we know that proprioception is sometimes impacted~\cite{niessen2008proprioception}. Without adequate sensing, interaction forces between limb segments become difficult to predict, disrupting inter-joint coordination~\cite{sainburg1993loss}. Secondly, authors in~\cite{chae2002delay, barker2009training} report that initiation and termination of muscle contraction are significantly delayed in the paretic arm, suggesting longer signal transmission from intent to action and leading to an inability to quickly change movement direction. A more recent study~\cite{nature-oscillatory} reports a significant difference in EEG activities during fast movements of non-paretic and paretic hands, implying that higher cognitive efforts could be required to perform fast repetitive movements in paretic limbs. These findings offer differing but non-contradictory perspectives---impaired motor coordination observable in post-stroke hemiparesis is likely caused by a combination of underlying neural mechanisms. 

\begin{figure*}[t]
  \centering
  \includegraphics[width=\textwidth]{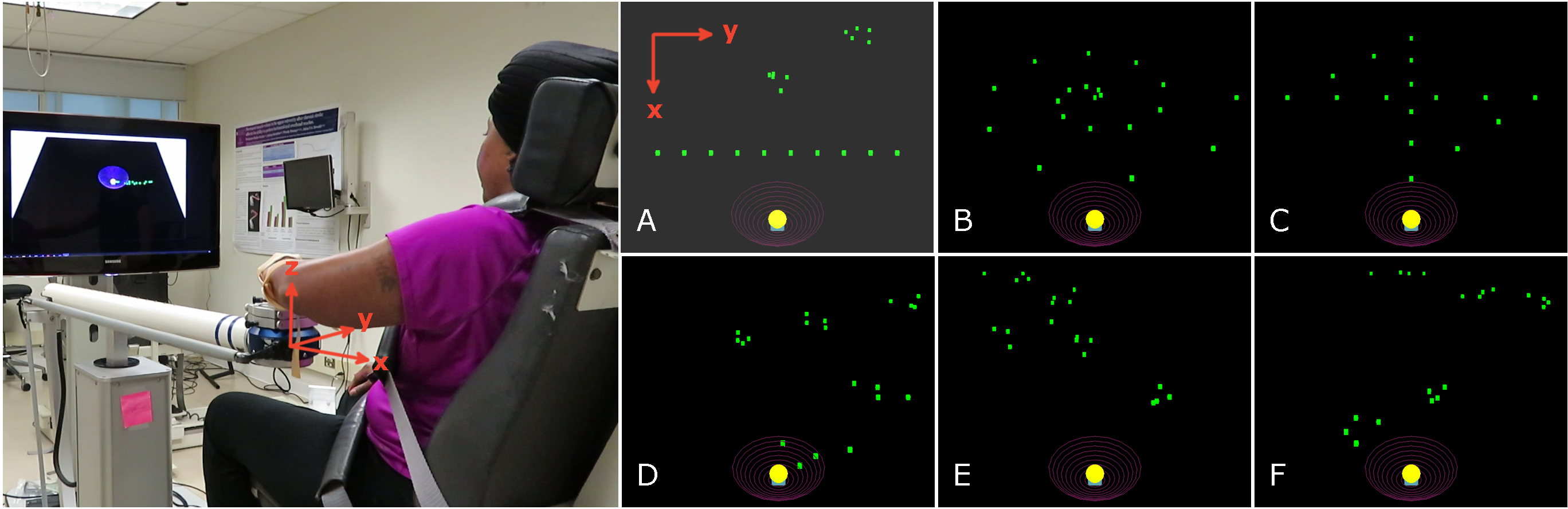}
  \caption{Experimental Setup. (left) Participant using the ACT\textsuperscript{3D} robot to complete the flag-collection task. (right) Task variations. A. Flag distribution used during learning phase. B-F. Five flag distributions used for data collection. }
  \label{fig: task_dist}
\end{figure*}

Our approach offers yet another perspective. Previous studies have found that abnormal muscular coupling resulting from a progressive increase in shoulder abduction loading impacts reach of motion and therefore reduces the size of an individual's reachable workspace~\cite{sukal2007shoulder}, presumably due to an increased use of indirect motor pathways~\cite{mcpherson2018progressive}. Here, we investigate whether abnormal muscular coupling and the use of these indirect pathways impacts timing-dependent motor coordination \textit{within} an individual's reachable workspace. We evaluate how performance in dynamic tasks---tasks that are highly timing-sensitive, such as carrying a glass of water or moving around a bowl with a ball rolling inside it---changes with different levels of shoulder abduction loading. In our experiments, we test this interdependence by evaluating individuals' performance at a virtual ball-in-bowl task with varying levels of abduction loading provided using an assistive robot with haptic feedback. 

\textbf{Our results show that there exists an interdependence between levels of abduction loading at the shoulder joint and impaired participants' ability to perform dynamic movements in their reachable workspace.} This is a novel insight---it indicates that there is a potential connection between the increased use of indirect multi-synaptic low resolution motor pathways in the paretic upper limb and timing-sensitive motor coordination within the reachable workspace. In our analysis, we observe that overall task performance, as measured by the average time needed to obtain a target, strongly depends on the load level provided. Moreover, impaired participants provide significantly smaller forces at the resonant frequency of the ball than do controls and this discrepancy increases with an increase in loading. Finally, we assess participants' quality of motion via their ability to make quick directional changes in movement and we see a similar statistical correlation---quality of movement decreases with increases in loading, particularly for individuals with chronic stroke. 

In the sections below, we describe our experimental design and analyses. We present experimental findings and conclude with a discussion of their implications and directions for future work. 

\section{Methods}

During experiments, participants are seated in front of a screen with their arm strapped to an upper-limb orthosis, as visible in Figure~\ref{fig: task_dist}. Once situated, they are asked to complete a virtual ball-in-bowl task with haptic feedback forces. Prior to the experiment, we assess each participant's maximum shoulder abduction (SABD) force and their reachable workspace---both parameters are later used to scale the difficulty of the experimental task. The experimental setup and procedure, as well as the ball-in-bowl task, are described in more detail below. In Section~\ref{sec: data analysis}, we describe metrics used to assess participants' performance and the statistical tests used during analysis. 

\subsection{Experimental Setup \& Virtual Task}
\label{sec: task}

The hardware used in this study is the Arm Coordination Training 3-Dimensional Device (ACT\textsuperscript{3D})~\cite{sukal2005dynamic, Sukal2007}, which integrates a HapticMASTER robot (MOOG, The Netherlands) with a Biodex experimental chair (Biodex Medical Systems, Shirley, NY) and replaces the end-effector with a rigid forearm-hand orthosis. During experiments, participants are seated in the Biodex chair with straps across their chest and lap to restrict movement of the trunk. The participant's arm is secured to the orthosis, allowing them to control the position of the end-effector with the movements of their arm. A virtual ball-in-bowl task is visualized using OpenGL on a screen in front of the participant with the location of the end-effector mapped to the location of a bowl on a virtual table visible on the screen. The robot's end-effector has 3 translational degrees of freedom that affect the game, while the orthosis can rotate in the xy plane for comfort but its rotation is not tracked throughout the experiment. 

The virtual task consists of manipulating the bowl with a ball rolling inside it while collecting flags. The ball-in-bowl system is adapted from \cite{Maurice2018}, where the system dynamics are simulated using a spherical cart-pendulum model. Motion of the bowl is limited to the horizontal plane, and the ball rolling inside is constrained to the end of a pendulum. As a result, the ball falling out of the bowl is not simulated and is instead indicated with a change in the color of the ball (from yellow to red). The HapticMASTER robot is used to provide haptic feedback corresponding to the interaction force between the bowl and the rolling ball. It is also used to render a haptic table to enable resting between task attempts. 

The length of the pendulum governing the motion of the ball is chosen based on the frequencies at which individuals with and without stroke can initiate movement. Barker et al. found that individuals with stroke begin to react to auditory stimuli with an average movement onset time of 0.81s for the tricep and 1.17s for the bicep, corresponding to frequencies of 1.23Hz and 0.85Hz respectively~\cite{barker2009training}. In contrast, the average movement onset time observed in this study for controls was 0.41s for the tricep and 0.25s for the bicep, corresponding to frequencies of 2.44Hz and 4Hz respectively. Therefore, for our experiment we chose a pendulum with a resonant frequency of 1.88Hz---greater than the average frequency at which individuals with stroke can respond to a stimulus and lower than the average capabilities of able-bodied controls. When navigating the virtual ball-in-bowl system, this choice of parameters makes the user feel as if they were moving a wooden bowl with a ping pong ball inside it. 

The goal of the virtual task is to collect as many flags as possible as quickly as possible within a 20 second time limit. Feedback of task performance is provided on the screen as the number of flags collected and the time remaining. Flags are displayed as dots on a virtual table. To be able to collect a flag, three criteria must be met:
\begin{itemize}
  \item The ball is inside the bowl.
  \item The participant's arm is lifted above the haptic table.
  \item The xy location of the bottom of the bowl matches the xy location of the flag within a small tolerance.
\end{itemize}
When the first two criteria are met, a blue square appears under the bowl to indicate that the participant is eligible to collect flags. When the third criterion is also met, the dot representing the flag disappears. 

Throughout the experiment, the virtual task is varied with respect to the distribution of the flags, later referred to as task variations, and the weight of the load, later referred to as loading levels. Six different distributions of 20 flags each, visible in Figure~\ref{fig: task_dist}, and three SABD loading levels are used. Of the six flag distributions, one is used for training and the remaining five are used for data collection. The loading levels are based on the maximum vertical SABD force. The lightest loading level, referred to as $0\%$, requires the participant to support the weight of their arm with no additional weight added by the HapticMASTER. The two heavier loading levels, $20\%$ and $50\%$, require the participant to support a percentage of their maximum vertical SABD force rendered by the HapticMASTER in addition to the weight of their arm. 

\subsection{Experimental Procedure}

Each participant's maximum SABD force is measured prior to task execution in an isometric setup, like the one in~\cite{isometric2007ellis, beer1999task}. Each participants' arm is positioned at 80\degree\ shoulder abduction, 90\degree\ elbow flexion, and 40\degree\ horizontal abduction. Once situated in the ACT\textsuperscript{3D}, participants are asked to reach out diagonally to the right and left under the maximum loading condition to create a conservative estimate of the participant's reachable workspace. The six flag distributions are scaled to fit within this estimated workspace (while preserving distribution shapes) to ensure that all flags are reachable under all three loading conditions.

Before data collection begins, participants are trained to manipulate the bowl and become familiar with the ball dynamics. Participants practice timed and untimed trials using a training flag distribution---illustrated in Figure~\ref{fig: task_dist}~A. Once participants demonstrate an understanding of how to perform the task, data collection begins.

\begin{figure}[t]
  \centering
  \includegraphics[width=\columnwidth]{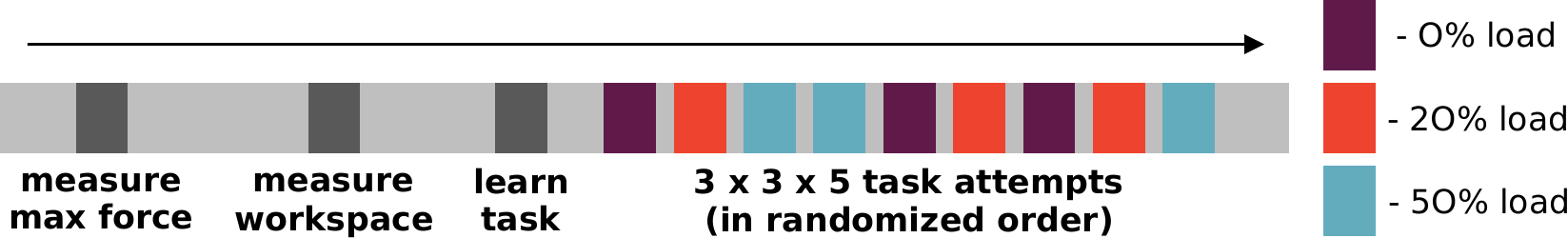}
  \caption{Experimental procedure with an example order of randomized load levels. Within each block of 5 task trials, participants completed all five different task variations. }
  \label{fig: exp-procedure}
\end{figure}

During data collection, participants perform the virtual task in a total of 45 trials. Trials are divided into nine sets of five trials each. Each set of trials consists of one instance of each flag distribution performed at a given loading level---three sets are performed at each loading level. The order of flag distributions within each set and the order of loading level for each set is randomized to account for fatigue and learning in our later analysis. Breaks are given as needed between each set of trials. The experimental procedure---with an example random order of loading levels---is shown in Figure~\ref{fig: exp-procedure}.

\subsection{Recruited Participants}
Six individuals (3 male and 3 female, 43 to 71 years of age) with chronic stroke (11-16 years post infarct) were recruited for the stroke group of this study. Of the participants with chronic stroke, 4 were left hemiplegic and 2 were right hemiplegic. Additionally, six able-bodied individuals (3 male and 3 female, 24 to 34 years of age) were recruited for the control group. All participants provided informed consent prior to taking part in the study, which was approved by the Institutional Review Board at Northwestern University (IRB STU00021840).

\begin{figure*}[t]
  \centering
  \begin{subfigure}%
    \centering
    \includegraphics[width=.49\textwidth]{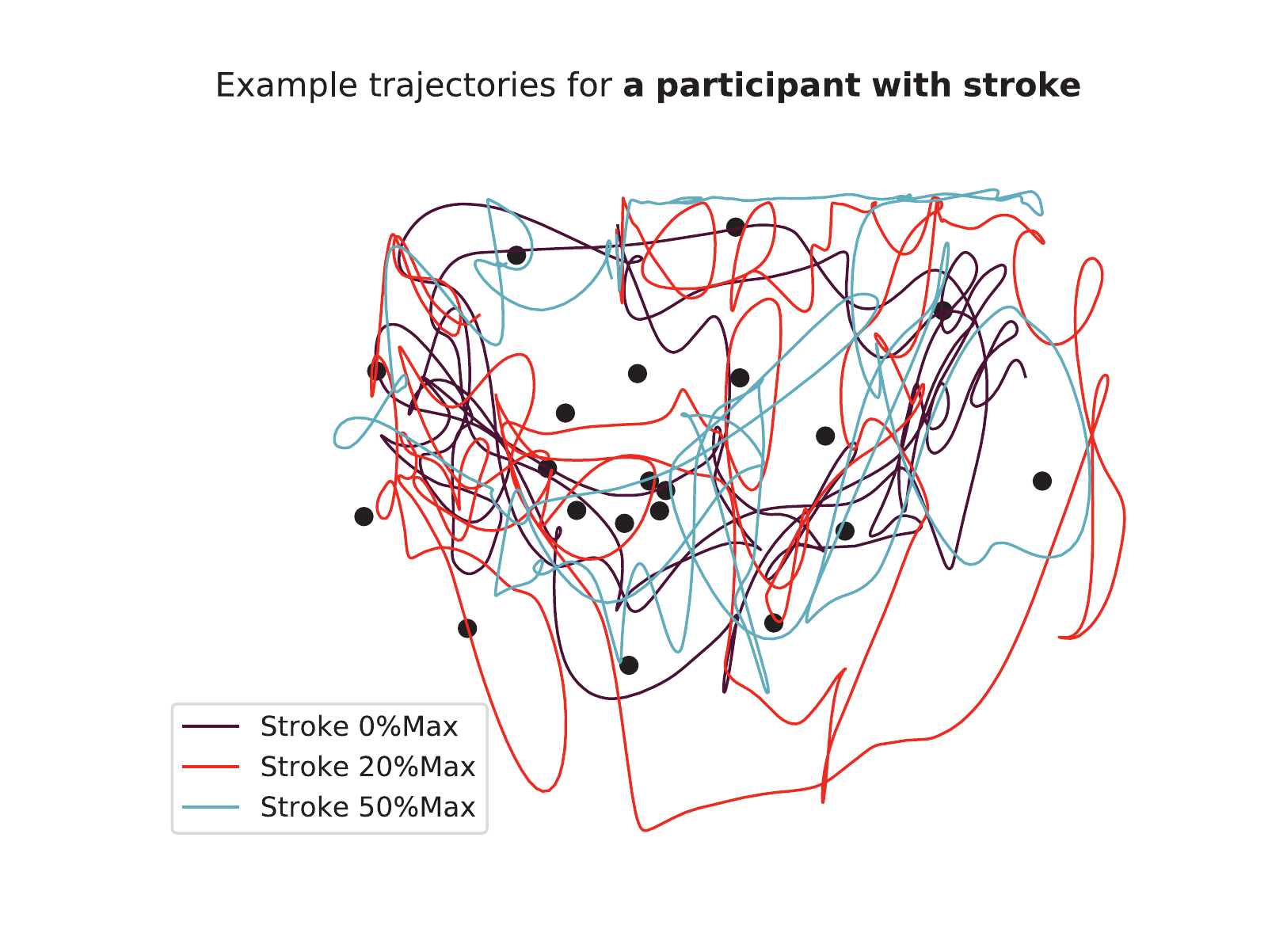}
  \end{subfigure}%
  \begin{subfigure}%
    \centering
    \includegraphics[width=.49\textwidth]{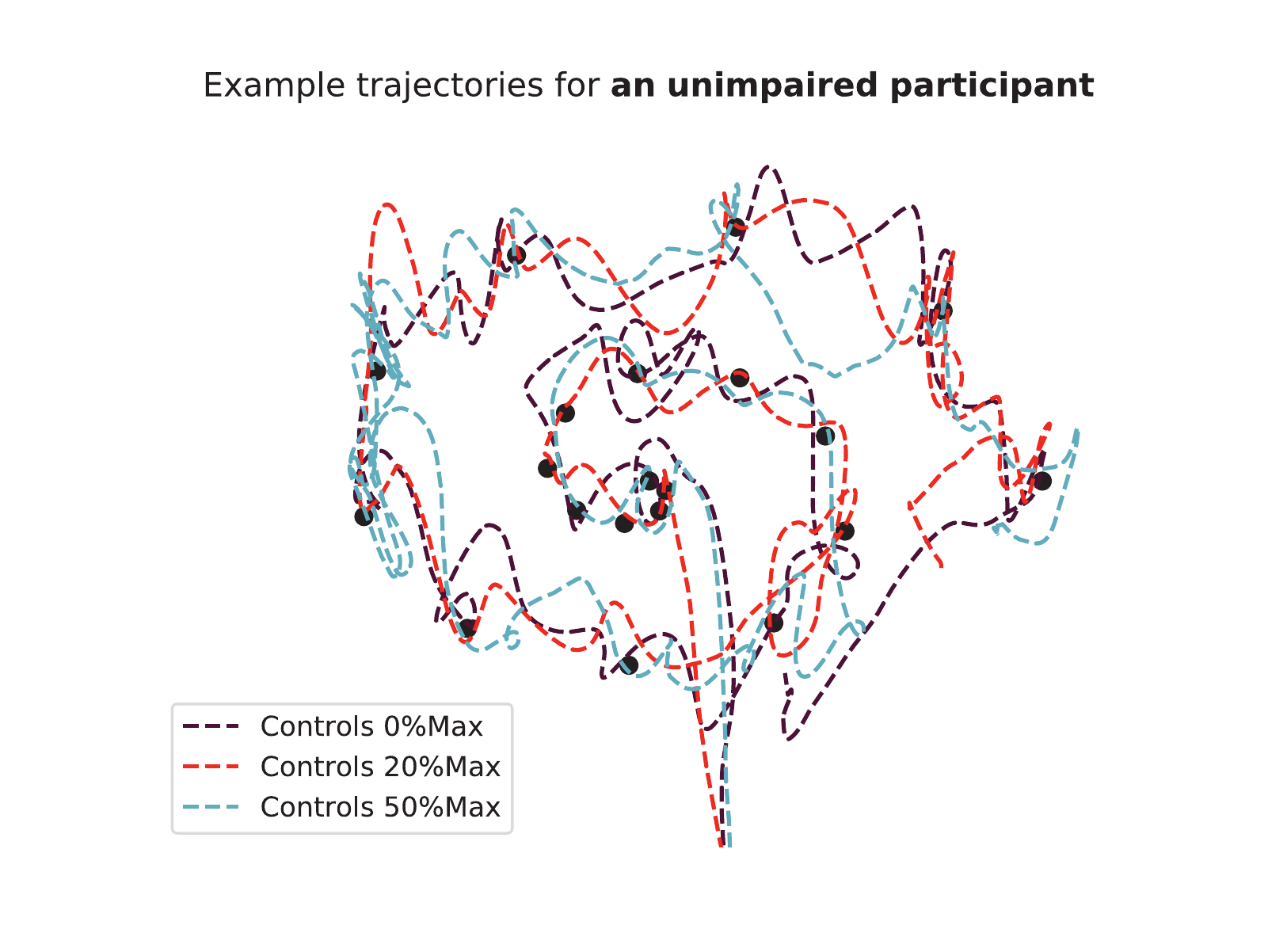}
  \end{subfigure}
  \caption{Example trajectories of movement for an impaired (left) and an unimpaired individual (right) for the same task distribution at all 3 loading levels. Note that we see an overall inability to precisely control movement for the impaired participant, which leads to many low frequency oscillations in their trajectories---this low-frequency activity is also visible in the frequency decomposition of forces.}
  \label{fig: trajectories}
\end{figure*}

\subsection{Metrics \& Performance Analysis}
\label{sec: data analysis}

During each task attempt, we record two time-specific metrics---the number of flags collected during that trial and the time spent performing the task (task-time). The task-time is counted from start until the last flag is collected or until the trial ends (20 seconds), whichever comes first, but only while the participant's arm is lifted above the haptic table (and supporting the prescribed load for that task attempt). Time while the participant rests on the haptic table is not counted towards the task-time for that attempt. Finally, we characterize overall task performance using a time-per-target metric, calculated by dividing task-time by the number of targets collected. 

During each attempt, we also collect time series data of forces exerted by the participant in the xy-plane. The ACT\textsuperscript{3D} robot is equipped with a 3-dimensional load cell that enables us to measure xyz forces applied at the end-effector. Front-and-back reaching movements are parallel to the x-axis; side-to-side movements are parallel to the y-axis; and up-and-down lifting movements are parallel to the z-axis, as indicated in Figure~\ref{fig: task_dist}. Note that the ball-in-bowl task is highly dynamic because of the ping-pong-like ball that is able to roll around inside the bowl. When energy is introduced into the ball-in-bowl system through movement, the system becomes dominated by the resonant frequency of the ball. The ball's oscillations must be counteracted to successfully damp out energy, control the ball's activity, and achieve optimal task performance. As such, we evaluate the frequency decomposition of participants' motion during individual task attempts.

Spectral analyses are performed on the x and y components of force. A discrete fast Fourier transform (FFT) is performed on each trial to obtain the amplitude of the signal for a range of frequencies up to the Nyquist frequency of 50Hz for the 100Hz sampling rate. In a frequency spectrum, the amplitude of a signal at a certain frequency represents the relative energy of the signal at that frequency, providing us with insight about participants' ability to switch movement direction. To enable direct trial to trial comparison, the signals are normalized by trial length. Even so, signals vary in total energy due to variation in participants' maximum strength and workspace area. To allow for comparison between participants, the signal in the frequency domain is also normalized by the energy introduced into the ball-in-bowl system by a participant throughout a task attempt. After the normalization, the total energy introduced into the system during each task attempt is equal to one, allowing us to do trial to trial and subject to subject comparisons. For each group and loading level, the frequency decompositions are averaged across participants and presented in Figure~\ref{fig: frequency}. It is worth noting that the trials range from 5.2 to 20 seconds in length, allowing us to obtain a 0.05-0.2Hz resolution in the frequency domain. 

Four metrics are obtained to evaluate performance in the frequency domain: `high/low frequency ratio' and the `peak amplitude near resonance', both in the x-direction and y-direction of the force signal. The `peak amplitude near resonance' is estimated as the maximum signal energy within $\pm$1Hz of the resonant frequency. The `high/low frequency ratio' is calculated by dividing the signal energy at high frequencies ($\geq$1Hz) by the signal energy at low frequencies ($<$1Hz). A cut-off value of 1Hz is chosen because it is in between the average frequency in which the biceps and triceps muscles can react to stimuli in individuals with stroke, as mentioned in Section~\ref{sec: task}. For each metric, a mixed-design ANOVA between groups (individuals with and without stroke) with within-participant factors for loading level and task is performed in R ($\alpha=0.05$). For each group and metric, a 2-way repeated measures ANOVA for loading level and task is performed ($\alpha=0.05$). Sphericity is evaluated using Mauchly's sphericity test, and if the sphericity assumption is violated, the p-value associated with the Greenhouse-Geisser correction is provided.

\section{Results}
\label{sec: results}

Participants' overall performance is evaluated using task-specific metrics, such as time-per-target. Their ability to control the dynamics of the ball is assessed by comparing the frequency spectra of their motion. Our results show that abduction loading impacts timing-sensitive motor coordination---including overall task performance and ability to react to the motion of the ball---significantly more for individuals with a hemiparetic stroke than for able-bodied controls. Example task executions for an impaired and non-impaired individual are visualized in Figure~\ref{fig: trajectories}.

\subsection{Overall task performance} 
\label{sec: overallperformance}

\begin{figure}[t]
  \centering
  \includegraphics[width=\columnwidth]{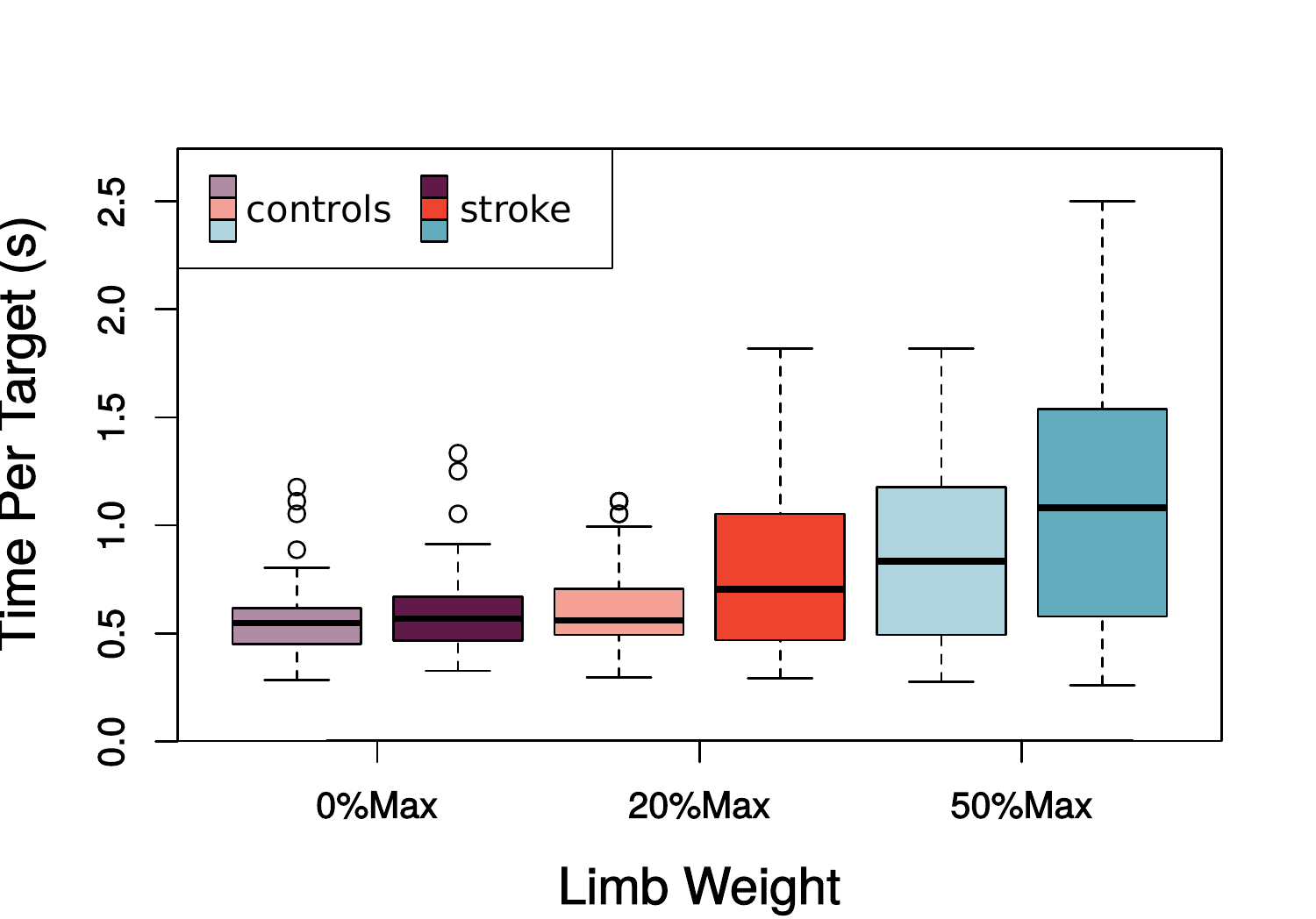}
  \caption{Average performance---as measured using time per target---across both groups at three different load levels (0\%, 20\%, and 50\%). Notice that performance changes more significantly with loading for the stroke group than for controls. }
  \label{fig: time-per-target}
\end{figure}

We use a repeated-measures ANOVA to test for the effect of loading on overall task performance within the two groups of participants. SABD loading is a significant factor for time-per-target for stroke participants (F=3.7,~p=0.06), while it is not a significant factor for able-bodied individuals (F=2.06,~p=0.18). When both groups are compared together in a mixed-design ANOVA with group (stroke or control) treated as another factor, both group (F=3.95,~p=0.08) and SABD loading level (F=4.7,~p=0.02) are significant factors. Moreover, there is an interaction effect between group and loading level (F=2.55,~p=0.10), indicating that SABD loading impacts time-per-target differently for stroke survivors than it does for healthy controls. Average performance for the two groups at three loading levels is visualized in Figure~\ref{fig: time-per-target}.

\subsection{Response to ball dynamics}
\label{sec: frequency}
\begin{figure*}[t]
  \centering
  \includegraphics[width=\textwidth, height=12cm, keepaspectratio]{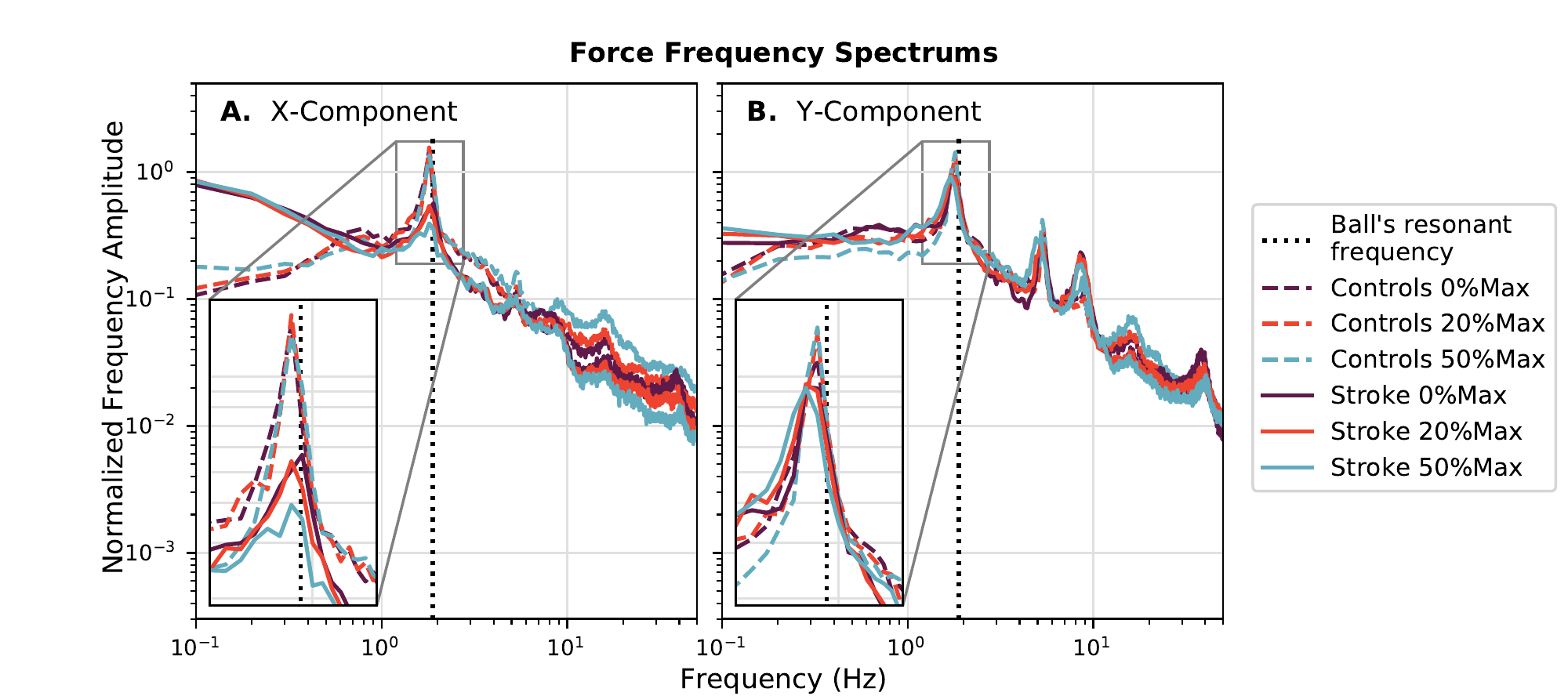}
  \caption{Averaged frequency spectra in the x- and y-components of force for all 12 individuals with and without stroke, calculated based on all trials. The amplitude is normalized by signal energy, as explained in Section~\ref{sec: data analysis}; a zoomed-in section of the plots is provided around the resonant frequency of the ball (1.88$\pm$1Hz) to better illustrate the differences in peaks near resonance. Note that although all participants exert significant energy counteracting the ball's oscillations, able-bodied controls do it on average more. Moreover, the ability to generate these forces decreases with loading. It decreases more for individuals with stroke than for able-bodied controls.}
  \label{fig: frequency}
\end{figure*}

In Section~\ref{sec: overallperformance}, we show that overall task performance differs according to group and experimental conditions. Here, we analyze \textit{how} it differs by looking independently at one aspect of the task---the ability of participants to counteract ball dynamics. To succeed at the task---in addition to collecting dots---participants aim to prevent the ball from falling out of the bowl. In doing so, they respond to the ball's steady-state dynamics and provide forces that oscillate near the ball's resonant frequency, as visible in Figure~\ref{fig: frequency}. To compare participants' ability to make quick movements in response to the motion of the ball, we compare activity near the resonant frequency of the simulated ball. We use the metric `peak amplitude near resonance' for the x- and y-components of force described in detail in Section~\ref{sec: data analysis}. We calculate it for each task attempt and compare across group and experimental conditions.

In the mixed-design ANOVAs, group (individuals with stroke versus controls) is a statistically significant factor for both the x-component (F=41.8,~p=7.17E-5) and the y-component of force (F=9.49,~p=1.16E-2) applied near the ball's resonance, indicating that impairment has an impact on participants' ability to respond to the ball's dynamics. One potential explanation for these results is a difference in strategy employed by the two groups based on their physical capabilities. Oscillations of the ball can be damped in a variety of ways. One can aim to completely cancel out the oscillations of the ball by exerting an equal and opposite force to the haptic feedback. This strategy results in a peak in the force frequency spectrum at the ball's resonant frequency. We observe that individuals with stroke counteract less of the ball's resonant dynamics---they have a significantly smaller peak amplitude at the ball's resonant frequency. A likely explanation is that individuals with stroke have overall more difficulty modulating force at higher frequencies, because they rely more on lower resolution indirect motor pathways~\cite{mcpherson2018progressive}. Given the stroke-induced neural constraints, individuals with stroke respond to the dynamics by moving at lower frequencies, employing a strategy that involves moving the bowl steadily around the workspace in an attempt to keep the ball swirling near the bottom of the bowl. 

We again use ANOVAs to evaluate whether loading level affects the `peak amplitude near resonance' for both components of force. Loading level is a statistically significant factor for force in the x-direction for both groups combined (F=18.7,~p=2.68E-5) as well as independently in the repeated measures ANOVAs---for individuals with stroke (F=15.8,~p=8.11E-4) and controls (F=5.31,~p=2.68E-2). In the y-component, loading level is not a statistically significant factor in any of the ANOVAs (mixed-design (F=0.58,~p=5.67E-1), repeated-measures for individuals with stroke (F=1.12,~p=3.63E-1), or repeated-measures for controls (F=0.18,~p=8.39E-1)). Overall, when loading level is decreased, all participants exert more energy counteracting the ball's resonant frequency, likely because counteracting the ball's dynamics is more conducive to performing the task \textit{quickly} than other strategies. This effect is slightly more pronounced for individuals with stroke than for controls as demonstrated by greater statistical strength in the repeated measures ANOVAs and by visible differences in the aggregated results in Figure~\ref{fig: frequency}. Since differences in the `peak amplitude near resonance' metric according to loading level mirror that of the time-per-target metric, differences in overall performance are likely at least in part due to variation in the participants' ability to counteract the ball's dynamics under different experimental conditions.

\subsection{Quality of movement}
\label{sec: movement}
Individuals with stroke have difficulty initiating and terminating movement~\cite{chae2002delay, barker2009training} and they rely heavily on indirect low resolution motor pathways~\cite{mcpherson2018progressive}, which hinders moving at high frequencies. As mentioned above, a possible explanation for why individuals with stroke perform worse at timing-sensitive dynamic tasks is that they generally experience more difficulty moving at high frequencies. To evaluate this, we perform statistical tests on the `high/low frequency ratio' for the x- and y-components of force. The mixed-design ANOVA reveals group is a statistically significant factor for the x-component of force (F=35.6,~p=1.37E-4) with individuals with stroke having lower ratios compared to controls, but group is not a significant factor for the y-component of force (F=0.78,~p=3.98E-1). Loading level is a statistically significant factor for the x-component of force for both groups combined (F=11.0,~p=6.11E-4,~p[GG]=5.36E-3) as well as independently in the repeated measures ANOVAs---for individuals with stroke (F=7.69,~p=9.51E-3) and controls (F=17.5,~p=5.39E-4). For the y-component of force, loading level is a statistically significant factor for both groups combined (F=3.96,~p=3.55E-2) and the repeated-measures ANOVA for individuals with stroke (F=8.55,~p=6.84E-3), but loading level is not a statistically significant factor in the repeated-measures ANOVA for controls (F=0.15,~p=0.86). 

\begin{figure*}[t]
  \centering
  \includegraphics[width=\textwidth]{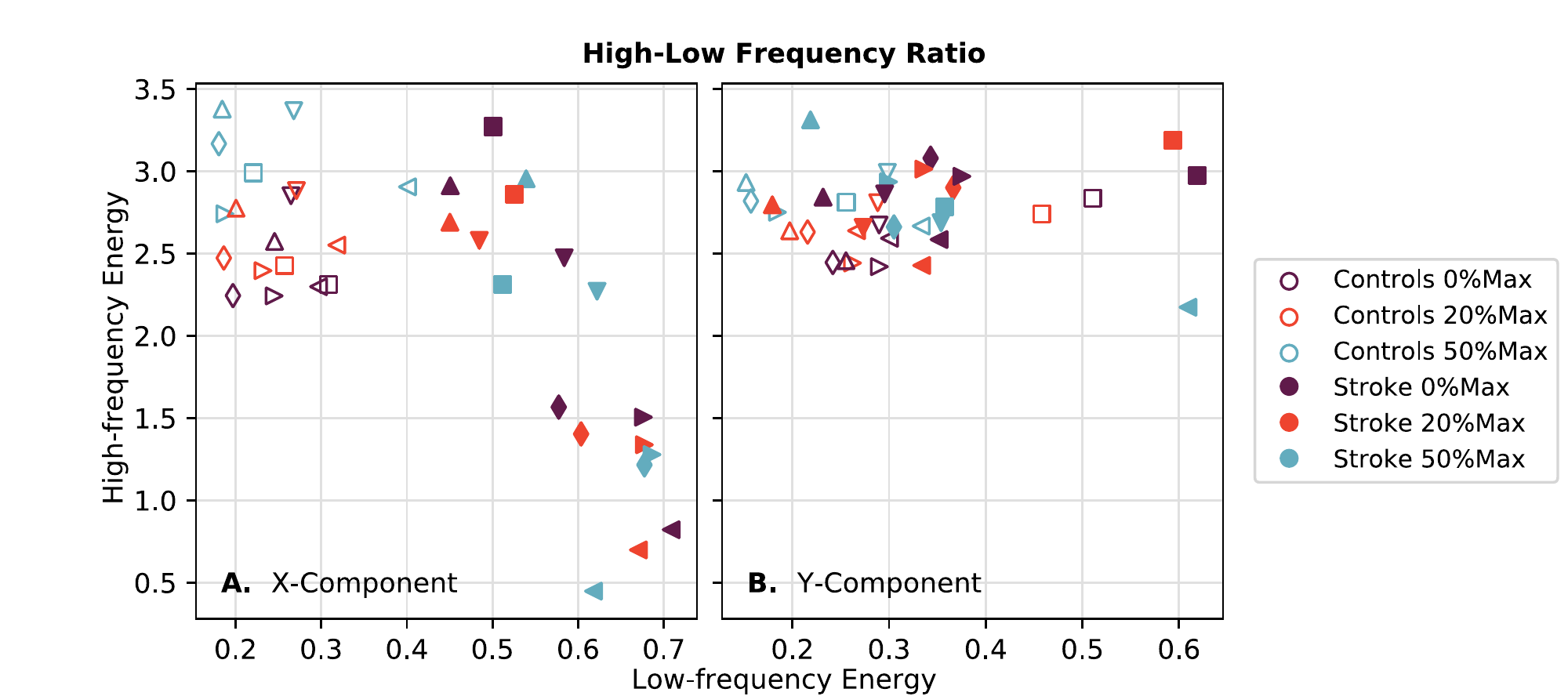}
  \caption{A visualisation of participants' dominant movement frequencies. Each shape represents one individual participant. Note, there exist visible differences in the average movement frequency between individuals with stroke and controls---controls move at higher frequencies. At the same time, note the impact of loading on the relative high/low frequencies of movement for individual participants---on average, loading has more impact on individuals with stroke than it does on healthy controls.}
  \label{fig: freqratio}
\end{figure*}

Importantly, there is a statistically significant interaction effect in the mixed-design ANOVA between group and loading level for the x-component of force (F=22.7,~p=7.18E-7,~p[GG]=3.79E-4) and the y-component of force (F=5.86,~p=9.91E-3,~p[GG]=2.72E-2), indicating that loading level affects the `high/low frequency ratio' differently for the two groups. For individuals with stroke, the `high/low frequency ratio' decreases with loading level, and for controls, the `high/low frequency ratio' increases with loading level. Overall, these results suggest that individuals with stroke experience more difficulty than controls moving at higher frequencies and the effect is exacerbated by SABD loading. This is likely due to a loss of independent joint control and a reliance on indirect low resolution motor pathways, both of which increase significantly in the presence of shoulder abduction loading~\cite{dewald1995abnormal,mcpherson2018progressive}. A visual representation of these results is provided in Figure~\ref{fig: freqratio}. 

\section{Discussion and Future Work}
In this section, we discuss the implications of our work and describe how it can be further developed. We explain how our methods can be used to create quantitative metrics of dynamic upper-limb motion and suggest how our results can help design robotic rehabilitation protocols for individuals with chronic hemiparetic stroke. 

\subsection{Assessment of timing-sensitive motor skills}
\label{sec: taskvariation discussion}
Objective quantitative metrics of motion are greatly beneficial in clinical practice---they allow for quick and consistent assessment of impairment, for monitoring recovery, as well as for generating robot-assisted training protocols. However, it is nontrivial to create these metrics, particularly for timing-sensitive motor coordination. Some quantitative metrics of motion exist and are already used to assess upper-limb motion post stroke, for instance:   
\begin{itemize}
    \item reachable workspace area~\cite{Ellis2011},
    \item smoothness of motion (as measured by jerk)~\cite{neville2002movementmetrics, bartolo2014progressive},
    \item circle drawing performance, where circle roundness and area are calculated~\cite{circledrawing, circledrawing2},
    \item error from a pre-determined trajectory~\cite{vergaro2010trackingerror},
    \item ergodicity of motion~\cite{fitzsimons2019ergodicity},
    \item or delays in movement initiation~\cite{chae2002delay} 
\end{itemize}
with error during trajectory-tracking and delays in movement initiation being most suitable for assessing timing-sensitive motion. However, these metrics allow us to assess performance only in pre-determined, often quasi-dynamic movements, like reaching to a target---they do not allow testing of dynamic responses without pre-specifying movement trajectories. As described in Sections~\ref{sec: data analysis} and~\ref{sec: frequency}, in this paper we offer a method for assessing the ability to generate fast coordinated movements while giving participants' freedom to choose their own movement strategy. This allows participants to employ compensatory movement strategies that they have developed post injury and would also employ during activities of daily living. 

One possible application of this metric that we have not yet explored is to quantify motor coordination in different areas of a reachable workspace. As of recently, successful training protocols exist for expanding reachable area post-stroke~\cite{Ellis2009progressive,Sukal2007,bartolo2014progressive}. However, functional ability in newly reachable distal regions has not been assessed. It is worth noting that ability to move to a given position becomes less valuable if these individuals are unable to perform meaningful tasks at that position (e.g. ability to reach a shelf means little if one cannot pick up an object from the shelf). Our approach could be used to assess performance in different sections of the reachable workspace. 

Interestingly, our results already indicate that there is variability in performance with changes in target distributions. As an example, we find that task variations---corresponding to different flag distributions---are a significant factor in the participants' performance. For instance, we observe interaction effects between group and task variation for the `peak amplitude near resonance’ metric for both x- (F=8.81, p=3.39E-5) and y-components of force (F=3.04, p=2.79E-2). This means that depending on the location of targets in participants' workspace, they show on average different ability to compensate for ball dynamics. More analysis would have to be done to explicitly evaluate location-dependent performance. In future experiments, we could design task variations to target assessment in specific areas of their workspace. Such follow-up work could lead to the development of a useful metric that is complementary to the currently used reachable workspace area. 

\subsection{Progressive shoulder abduction loading in stroke rehabilitation}
\label{discussion: loading rehab}
Progressive shoulder abduction interventions and gravity-compensating orthoses have shown promise for expanding reachable area post-stroke. In~\cite{Ellis2009progressive}, the experimental group that underwent training with progressive shoulder abduction loading experienced larger increases in workspace area than did individuals with stroke who trained for an equivalent amount of time without abduction loading, while no changes in strength were found. In~\cite{orthosiseffect}, subjects reached toward static targets at a self-selected speed with and without a compensating-for-gravity orthosis. Training with the orthosis led to significant decreases in jerk (improved quality of movement) and significant shifts in timing of peak speed (faster ability to initiate movement). Other studies~\cite{circledrawing, circledrawing2} investigated whether robotic gravity compensation training has an effect on unassisted circle drawing. Results showed that training improved circle area (functional workspace) with no effect on circle roundness (synergies persisted). All in all, these results indicate that progressive shoulder abduction loading can be utilized to successfully improve reaching range of motion. This is enabled by neural plasticity in the body and the body's ability to progressively activate remaining high resolution motor pathways from the lesioned brain hemisphere through training~\cite{wilkins2017neural}. 

Finally, it is worth noting that these therapies were initially investigated because it was known that reaching area varied with levels of neural drive of shoulder abductors. Given our results and experimental setup, we can design progressive abduction loading therapy focused on increasing upper limb coordination post stroke. We can achieve this presumably by reducing the maladaptive dependence on indirect motor pathways which in turn reduces dynamic response frequencies, as demonstrated in this work. In fact, recent neural modeling work~\cite{sinha2020cross} provides a theoretical framework for the benefits of relying more on these indirect motor pathways, as occurs during progressive shoulder abduction loading. Future work should investigate whether shoulder abduction loading can successfully augment dynamic task training. Moreover, we can enhance the dynamic task training through real-time feedback, which has been shown to accelerate learning in a non-impaired population~\cite{RSS2018-sharedcontrol}.

\section{Conclusions}

Our results show that there exists an interdependence between abduction loading at the shoulder joint and impaired participants' ability to perform dynamic movements in their reachable workspace. Given this finding, we can design targeted robot-assisted rehabilitation protocols focused on increasing overall limb coordination for chronic survivors of stroke. Increased arm coordination could significantly improve patients’ independence and quality of life. 

\addtolength{\textheight}{-0cm}  


\section*{ACKNOWLEDGMENT}

This material is based upon work supported by the National Science Foundation Graduate Research Fellowship Program under Grant No. DGE-1842165, by the National Science Foundation (NSF) under Grant No. 1637764, by the National Defense Science and Engineering Graduate (NDSEG) Fellowship program, and by NIH R01 grants NS105759 and HD039343 to JPAD. Any opinions, findings, and conclusions or recommendations expressed in this material are those of the authors and do not necessarily reflect the views of the NSF or NDSEG program.

The authors would like to thank Jasjit Deol, PT for her help with data collection and support throughout the project.


\bibliography{references}
\bibliographystyle{plain}

\end{document}